\documentclass[letterpaper, 10 pt, conference]{ieeeconf}  

\IEEEoverridecommandlockouts                              

\usepackage{amsmath} 
\usepackage{amssymb}  
\usepackage{graphicx}
\usepackage{booktabs}
\usepackage{colortbl} 
\usepackage[dvipsnames]{xcolor}
\usepackage{bm}
\usepackage{subcaption}

\definecolor{TableDarkGreen}{RGB}{182,215,168}
\definecolor{TableLightGreen}{RGB}{207,234,215}
\definecolor{TableYellow}{RGB}{255,242,204}
\definecolor{TableRed}{RGB}{244,204,204}

\usepackage{multirow, makecell}
\usepackage{arydshln}
\usepackage{hhline}
\usepackage[most]{tcolorbox} 
\usepackage[export]{adjustbox}

\title{\LARGE \bf
A-SCoRe: Attention-based Scene Coordinate Regression for wide-ranging scenarios
}

\author{Huy-Hoang Bui$^{1}$, Bach-Thuan Bui$^{1}$, Quang-Vinh Tran$^{1}$, Yasuyuki Fujii$^{2}$, Dinh-Tuan Tran$^{2}$, and Joo-Ho Lee$^{2}$
\thanks{*This work was not supported by any organization}
\thanks{$^{1}$Huy-Hoang Bui, Bach-Thuan Bui and Quan-Vinh Tran is with Graduate School of Information Science and Engineering, Ritsumeikan University, Osaka, Japan.}%
\thanks{$^{2}$Yasuyuki Fujii, Dinh-Tuan Tran and Joo-Ho Lee with the College of Information Science and Engineering, Ritsumeikan University, Osaka, Japan.}%
}

\begin{document}

\maketitle
\thispagestyle{empty}
\pagestyle{empty}

\begin{abstract}


Visual localization is considered to be one of the crucial parts in many robotic and vision systems. While state-of-the art methods that relies on feature matching have proven to be accurate for visual localization, its requirements for storage and compute are burdens. Scene coordinate regression (SCR) is an alternative approach that remove the barrier for storage by learning to map 2D pixels to 3D scene coordinates. Most popular SCR use Convolutional Neural Network (CNN) to extract 2D descriptor, which we would argue that it miss the spatial relationship between pixels. Inspired by the success of vision transformer architecture, we present a new SCR architecture, called A-ScoRe, an Attention-based model which leverage attention on descriptor map level to produce meaningful and high-semantic 2D descriptors. Since the operation is performed on descriptor map, our model can work with multiple data modality whether it is a dense or sparse from depth-map, SLAM to Structure-from-Motion (SfM). This versatility allows A-SCoRe to operate in different kind of environments, conditions and achieve the level of flexibility that is important for mobile robots. Results show our methods achieve comparable performance with State-of-the-art methods on multiple benchmark while being light-weighted and much more flexible. Code and pre-trained models are public in our repository: https://github.com/ais-lab/A-SCoRe.

\end{abstract}

\section{INTRODUCTION}






Visual localization refers to the task of estimating accurately the 6-DoF absolute pose including the rotation and translation of a camera, given an image. Due to its necessity, visual localization is a core component for many applications ranging from robot, self-driving cars to Augmented/Virtual Reality (AR/VR).

Many visual (re)localization methods have been proposed. However, they are often limited by certain constraints or lack versatility for diverse situations in robotic operational environments. These approaches can be broadly categorized into image retrieval, feature matching-based methods, Absolute Pose Regression (APR), and Scene Coordinate Regression (SCR).

Image retrieval \cite{arandjelovic2016netvlad} is relatively simple and can scale to large environments. However, storing a database of images increases the storage burden. A similar issue arises with image-matching-based approaches \cite{sarlin2019coarse}. Although they achieve high accuracy by establishing 2D-3D correspondences through 2D-2D matching \cite{sarlin2020superglue}, they require substantial storage space. The 3D scene model and 3D features, obtained from SLAM \cite{mur2015orb} or Structure-from-Motion \cite{schonberger2016structure}, can take up several gigabytes. Moreover, storing images or 3D structures could expose sensitive information to outsiders, raising serious privacy concerns.

\begin{figure}[t]  
    \centering
    \includegraphics[width=0.95\columnwidth]{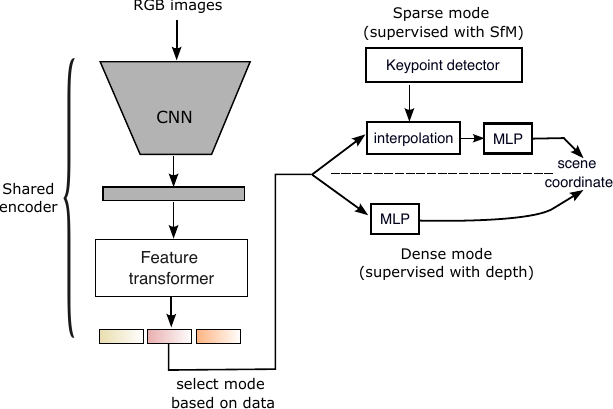} 
    \caption{Proposed method A-SCoRe overal architecture}
    \label{summary}
\end{figure}

Absolute Pose Regression (APR) and Scene Coordinate Regression (SCR) aim to address the limitations of previous approaches by leveraging deep learning. APR \cite{kendall2015posenet, bach2022featloc, wang2020atloc} learns a direct mapping from an image to its absolute pose, while SCR \cite{shotton2013scene, brachmann2017dsac, brachmann2018learning, brachmann2021visual, brachmann2023accelerated, bui2024d2s} takes a different approach by regressing the 3D scene coordinates of 2D pixels. Although APR is robust in terms of storage and speed, its lower accuracy makes it difficult to use in critical situations. On the other hand, once an SCR model accurately establishes 2D-3D correspondences, the camera pose can be reliably estimated using PnP+RANSAC \cite{kukelova2013real}. Methods such as DSAC \cite{brachmann2017dsac}, D2S \cite{bui2024d2s}, and ACE \cite{brachmann2023accelerated} have demonstrated exceptional performance, not only in accuracy but also in storage efficiency and computational cost.

By examining current SCR methods, we can broadly categorize them into two branches: dense prediction \cite{shotton2013scene, brachmann2017dsac, brachmann2021visual, li2020hierarchical} and sparse prediction \cite{bui2024d2s, do2022learning, tang2023neumap}. Although SCR performs well on multiple benchmarks, certain drawbacks limit its applicability in wide-ranging scenarios, especially for mobile robots.

Although both branches operate on the same principle, they differ significantly in architecture, ground-truth derivation, and training objectives. The dense branch typically uses a CNN to downscale the image, regress scene coordinates from the extracted feature map, and is supervised with a depth map. While CNNs are widely used, the recent success of Vision Transformers \cite{dosovitskiy2020image} has shown their ability to extract richer semantic information, leading to improved performance. Recent works in the sparse branch of SCR \cite{bui2024d2s, tang2023neumap} further support the use of attention mechanisms for visual localization. Additionally, \cite{nguyen2024focustune} highlights that dense prediction can lead to incorrect estimates for irrelevant parts of the scene, such as the sky or moving objects. 

\begin{figure}[t]  
    \centering
    \includegraphics[width=0.85\columnwidth]{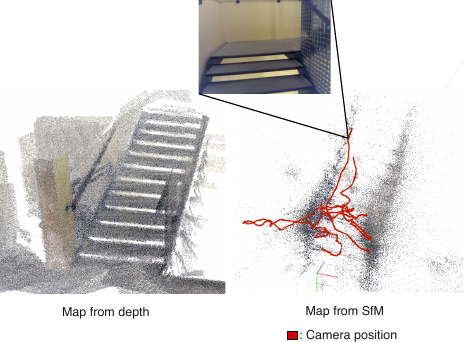} 
    \caption{Comparison of map built from depth and SfM. It is clear that sparse SfM omit much information where number of samples is low (upper part of the stairs). The image show a case of textureless region which cause difficulty for sparse SCR approaches.}
    \label{depth_vs_sfm}
\end{figure}

On the other hand, sparse methods benefit from existing photogrammetry algorithms to reconstruct sparse 3D scene models. Notably, Bui et al. \cite{bui2024d2s} proposed a model called D2S, which learns to map keypoint descriptors from a keypoint detector to Structure-from-Motion (SfM) scene coordinates. By leveraging powerful detectors like SuperPoint \cite{detone2018superpoint}, D2S focuses solely on keypoints rather than the entire scene, achieving superior performance in outdoor environments compared to dense approaches. However, the reliance on SfM also presents a weakness for methods like D2S, as the reconstructed 3D models lack information in textureless environments, as illustrated in Fig. \ref{depth_vs_sfm}.

Inspired by previous work on SCRs, we saw an opportunity to bridge the gap between different approaches. Thus, we propose A-SCoRe, a modular attention-based SCR that leverages the transformer architecture to enhance descriptors with spatial and semantic information. Additionally, our method is compatible with both dense depth maps and sparse SfM models, making it adaptable for robots equipped with either depth or RGB sensors.

In summary, our contributions are as follows:
\begin{enumerate}
    \item We propose A-SCoRe, a new SCR architecture that utilizes an attention mechanism on the descriptor map to enhance the semantic quality of 2D descriptors.
    \item A-SCoRe's design allows it to handle multiple input and output data modalities, making it significantly more versatile than existing SCR approaches.
    \item Extensive experiments demonstrate that the proposed method achieves performance comparable to state-of-the-art methods while remaining lightweight.
\end{enumerate}

\section{RELATED WORKS}


\textbf{Image retrieval} \cite{arandjelovic2016netvlad, torii201524, humenberger2022investigating} is the simplest form of visual localization in which a database of images captured the scene was kept. Whenever there is a query image, image retrieval system will compare similarity of the query with the ones in database to derive the camera poses \cite{camposeco2019hybrid}. This approach relies on how well can an image be represented as a feature vector. Methods such as NetVLAD \cite{arandjelovic2016netvlad} and its modified subsequent works \cite{humenberger2022investigating} were very popular. Recent image retrieval approach also utilize self attention and cross attention \cite{ali2024boq} in their model, which leads to improvements.

\textbf{Feature-matching-based approaches} \cite{sarlin2019coarse, humenberger2020robust, panek2022meshloc} estimate the camera pose by establishing 2D-3D correspondences between keypoints in images and scene coordinates in a 3D model. This 3D model is typically constructed using Structure-from-Motion (SfM) algorithms, notably COLMAP \cite{schonberger2016structure}. Since these approaches require a 3D map at runtime, they demand significant storage. Additionally, classical SfM can take a substantial amount of time for mapping, making it difficult to use in dynamic environments. Recent works have addressed these challenges. GoMatch \cite{zhou2022geometry} replaces descriptor-based matching with geometry-only matching, significantly reducing storage requirements. GLOMAP \cite{pan2024glomap} replaces COLMAP's incremental SfM with global SfM, achieving a major improvement in speed while maintaining the same level of accuracy. 


\textbf{Absolute Pose Regression (APR)} \cite{kendall2015posenet, wang2020atloc, bach2022featloc} directly regresses camera poses from images or predicts relative poses based on the top retrieved images \cite{turkoglu2021visual}. By leveraging neural networks to encode the image-to-pose relationship, APR requires significantly less storage and is much faster than feature-matching-based approaches. However, as noted in \cite{sattler2019understanding, ng2021reassessing}, its accuracy is closer to that of image retrieval methods. Recent works have attempted to improve APR by incorporating geometric constraints \cite{brahmbhatt2018geometry}, modeling uncertainty \cite{kendall2016modelling}, or utilizing outputs from SCR to refine predictions \cite{chen2024marepo}.


\textbf{Scene Coordinate Regression (SCR)} \cite{shotton2013scene, brachmann2017dsac, brachmann2021visual} predicts the 3D scene coordinates of 2D pixels, followed by PnP-RANSAC to compute absolute camera poses. Bui et al. \cite{bui2024d2s} adopted a similar strategy but focused on regressing only keypoints, proving to be more robust in environments affected by domain shifts, illumination changes, or extreme weather conditions. In \cite{li2020hierarchical}, the authors combined classification and regression for scene coordinate prediction. Instead of directly regressing scene coordinates, Do et al. \cite{do2022learning} proposed predicting the 2D projections of selected 3D points. SCR has also benefited from alternative representations, such as Neural Rendering \cite{10801953} and Gaussian Splatting \cite{zhai2024splatloc}, for improved performance. In subsequent work, Brachmann et al. \cite{brachmann2023accelerated} pretrained a feature extractor on a large dataset, enabling new scene mappings within minutes. The success of SCR has also contributed to foundational models in 3D reconstruction, such as ACEZero \cite{brachmann2024scene} and DUST3R \cite{wang2024dust3r}, which can reconstruct scenes from just two views, even with minimal or no overlap.


\begin{figure*}[t]  
    \centering
    \includegraphics[scale=1.2]{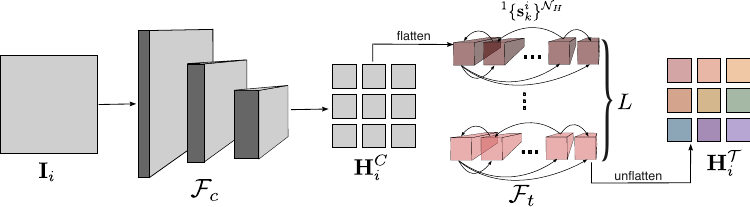} 
    \caption{\textbf{A-SCoRe shared image encoder}. From left to right, an image $\mathbf{I}_i$ pass through the convolutional layers denoted as $\mathcal{F}_c$.} 
    \label{backbone}
\end{figure*}

\section{METHODOLOGY}

Given a set of images $\{\mathbf{I}_i\}^{\mathcal{N}}$, where $i$ refer to the image index. Each image associated with one camera pose $\mathbf{T}_i$ and one intrinsic $\mathbf{K}_i$. $\mathbf{T}_i$ is a 3x4 matrix containing a rotation and translation component while $\mathbf{K}_i$ is a 3x3 matrix of the intrinsic calibration parameters. 

Our goal is to derive $\mathbf{T}_i$ using $\mathbf{I}_i$, which can be translated to the problem of estimating the 2D-3D correspondences. A pixel $j$ in image $\mathbf{I}_i$, denoted as $p^i_{j}$ can have a corresponding scene coordinate $\mathbf{x}^i_j$. Depends on methods and type of data that scene coordinate derivation are different, which are shown in subsequent sections.

\subsection{Image encoder}
This section describes the architecture of the image encoder shared across different data modalities, as illustrated in Fig. \ref{backbone}. The model comprises two main blocks: the feature extractor and the feature transformer, denoted as $\mathcal{F}_c$ and $\mathcal{F}_t$, respectively. When a RGB or RGB-D image $\mathbf{I}_i$ with size $(C, H, W)$ comes into the system, it will first pass through $\mathcal{F}_c$, which will produce a down-sized descriptor map:
\begin{equation}
    \mathbf{H}_i^C = \mathcal{F}_c (\mathbf{I}_i)
    \label{cnn}
\end{equation}
Descriptor map $\mathbf{H}_i^C$ resolution will be reduced while the channel depth will increase compared to the input image, resulting in a new shape $(D, \frac{H}{8}, \frac{W}{8})$. This reduction will allow us to perform operations with much less computation.

We associate a feature vector indexed $k$ of feature map $\mathbf{H}_i^C$ as $\mathbf{v}_k^i\in \mathbb{R}^D$, which corresponds to a patch in the image $\mathbf{I}_i$. After flattening, we obtain set of feature vector $\{\mathbf{v}_k^i\}^{\mathcal{N}_H}$, where $\mathcal{N}_H=\frac{H}{8}\times\frac{W}{8}$.

By passing the image through a transformation $\mathcal{F}_c$, we have extracted information in local region using the convolution operation. Next, we use transformer $\mathcal{F}_t$ to model the relationship between feature vectors, which results in a new attention feature map $\mathbf{H}_i^{\mathcal{T}}$.

\begin{equation}
    \mathbf{H}_i^{\mathcal{T}} = \mathcal{F}_t ({\mathbf{v}_k^i})
\end{equation}

Following inspiration from \cite{lindenberger2023lightglue}, we design the attention module $\mathcal{F}_t$ consists of $L$ identical layers. The state of vector $\mathbf{v}_k^i$ at layer $l$ denoted as ${}^{(l)}\mathbf{s}^i_k$. We update the state vector as it passing through each layer as follow:

\begin{equation}
    {}^{(l+1)}\mathbf{s}^i_k = {}^{(l)}\mathbf{s}^i_k + \sigma ([{}^{(l)}\mathbf{s}^i_k|\Psi({}^{(l)}\mathbf{s}^i_k, {}^{(l)}\mathbf{m}^i)] ) 
    \label{state_vector}
\end{equation}

where $[.|.]$ denotes the concatenation operation, $\sigma$ models the Multi-layer-Perceptron (MLP). ${}^{(l)}\mathbf{m}^i$ is the message coming from other feature vector of the same feature map $i$ and $\Psi$ is the multi-head attention, which is calculated as weighted average of all state vectors:

\begin{equation}
    \Psi({}^{(l)}\mathbf{s}^i_k, {}^{(l)}\mathbf{m}^i) = \sum_{k, n\in \mathcal{N}_H} \alpha_{kn}\cdot\mathbf{m}^i_n
    \label{attention}
\end{equation}

where $\mathbf{m}^i_n$ message is calculated as the linear projection from state vector $\mathbf{v}^i_n$, $\alpha_{kn}$ is the attention score of state vector $k$ and $n$, which is denoted in Eq. \ref{attention2}.
\begin{equation}
    \alpha_{kn} = {Softmax}_n(\mathbf{q}_k^T\cdot \mathbf{k}_n/\sqrt{D/h})
    \label{attention2}
\end{equation}

where $\mathbf{q}_k$, $\mathbf{k}_n$ denotes the queries, key vector calculated from linear projection of state vectors $\mathbf{v}^i_k$ and $\mathbf{v}^i_n$, and $\mathbf{h}$ is the number of heads.

After running through every layers of feature transformer $\mathcal{F}_t$, we obtain the final set of state vector ${}^{L}\{\mathbf{s}_k^i\}^{\mathcal{N}_H}$. Unflattening this set leaves us with the attention map $\mathbf{H}_i^{\mathcal{T}}$ with the same shape as $\mathbf{H}_i^C$. In subsequent steps, we will make prediction of scene coordinate based on this attention feature map.

\subsection{Dense mode}
We will first derive the ground truth coordinate for dense prediction. For each image $\mathbf{I}_i$, assume we have the corresponding depth measurement $\mathbf{D}_i$ and camera pose estimation from SLAM. We can calculate scene coordinate $\mathbf{X}_i$ by Eq. \ref{depth2sc}.

\begin{equation}
    \mathbf{X}_i = \mathbf{T}_i^{-1}(\mathbf{D}_i[\mathbf{K}_i, 1]^T)
    \label{depth2sc}
\end{equation}
where $\mathbf{T}_i^{-1}$ is the transformation from camera frame to scene coordinate frame system. Since the image after passing over $\mathcal{F}_c$ is downsized, we also downsize the scene coordinate map $\mathbf{X}_i$ by 8 times to obtain the final scene coordinate $\ddot{\mathbf{X}_i}$

\begin{figure}[t]  
    \centering
    \includegraphics[width=0.65\columnwidth]{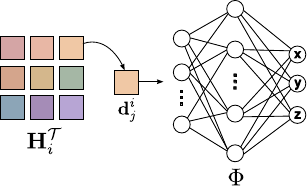} 
    \caption{\textbf{Illustration of the dense mode}. Each feature in the attention feature map will be map to a scene coordinate using the MLP network $\Phi$.}
    \label{dense_arch}
\end{figure}

The architecture for this mode is shown in Fig. \ref{dense_arch}, given the attention feature map, we utilize a MLP  to perform mapping from feature to scene coordinate. $\Phi: \mathbb{R}^D \rightarrow \mathbb{R}^3$
\begin{equation}
    \mathbf{x}_j^i = \Phi(\mathbf{d}_j^i)
\end{equation}

\subsection{Sparse mode}

In sparse mode, we rely on SfM algorithm \cite{schonberger2016structure} to reconstruct scene coordinate and camera pose. Since different keypoint detector will output different keypoint, we made an assumption that 3D model reconstructed using Scale-invariant feature transform (SIFT) features is the most reliable and camera poses from this model are ground-truth. 

Given camera poses from multiple views with known extrinsic and intrinsic, we can get the scene coordinate for the corresponding keypoint detector using triangulation. For each image $\mathbf{I}_i$, we first detect keypoints $p_j^i$ and then match those keypoints across $\mathcal{N}$ frames to obtain keypoint tracks $\mathbf{T}$. For each camera $i$, the projection is formulated as:
\begin{equation}
    \lambda_i p_j^i = \mathbf{K_i}\mathbf{T_i}\mathbf{x}_j^i
    \label{triangulation}
\end{equation}

where $\lambda_i$ is the depth scale factor. Initial estimation from Eq. \ref{triangulation} need to be optimize across all detected track by running bundle adjustment, which minimize the reprojection error:
\begin{equation}
    \sum_i ||p_j^i - \mathbf{K_i}\mathbf{T_i}\mathbf{x}_j^i||^2
\end{equation}

\begin{figure}[t]  
    \centering
    \includegraphics[width=1\columnwidth]{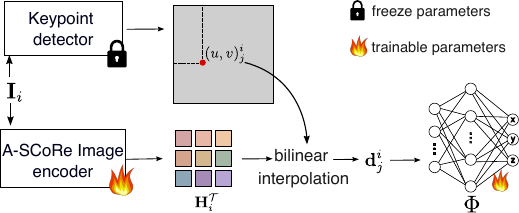} 
    \caption{\textbf{Illustration of the A-SCoRe sparse mode}. Keypoint detector output 2D pixel locations which are used to bilinear sample descriptor from attention feature map. Keypoint detector, if contain trainable parameters, are frozen.}
    \label{arch_sparse}
\end{figure}

After obtaining ground-truth scene coordinate, we will make sparse prediction based on the attention feature map. Fig. \ref{arch_sparse} illustrate this step. 

During training, any part of the keypoint detector is freezed, we only use it to extract a set of keypoint $\{p^i\}$ with coordinate $\{(u, v)^i\}$ of image $\mathbf{I}_i$. Since $\mathbf{H}_i^{\mathcal{T}}$ is downsized, we will first normalize the keypoints coordinate in range $[-1, 1]$ to get $\{u_n, v_n)^i\}$, then compute the bi-linear interpolation to obtain a set of keypoint descriptor $\{\mathbf{d}^i\}$ using Eq \ref{bi-linear}. 

\begin{equation}
\begin{aligned}
    \mathbf{d}_j^i = \sum_{a=0}^1 \sum_{b=0}^1 w_{ab}\cdot \mathbf{H}_i^{\mathcal{T}}(y_b, x_a) \\
    w_{ab}=(1-\mathbf{du})^{1-a}\cdot \mathbf{du}^a\cdot (1-\mathbf{dv})^{1-b}.\mathbf{dv}^b
\end{aligned}
\label{bi-linear}
\end{equation}

where $w_{ab}$ is the bilinear weights, $\mathbf{du}=u_n - u_0$ with $u_0$ is the floor indices of the nearest pixel, $\mathbf{dv}$ can be calculated in similar manner.

At this stage, it becomes similar to the dense mode, we apply MLP $\Phi$ on the descriptor to obtain the scene coordinate of detected keypoints.

\subsection{Supervision}
At training time, for the dense mode, we will learn the weights of the image encoder and the mapping MLP $\Phi$. For the sparse mode, the additional keypoints extractor is freezed, the rest will be learnable like the dense mode. Since in both cases, we have the ground-truth scene coordinate, we train the model in supervised manner by minimizing loss $\mathcal{L}$ using the L2 norm as in Eq. \ref{loss}.

\begin{equation}
    \mathcal{L} = \sum_i \sum_j ||\mathbf{x}_j^i - \hat{\mathbf{x}}_j^i||^2
    \label{loss}
\end{equation}

where $\hat{\mathbf{x}}_j^i$ is the scene coordinate prediction from A-SCoRe model.

\section{EXPERIMENTS}

\begin{table*}
  
\caption{\textbf{Indoor localization results on 7Scenes} \cite{shotton2013scene}. Median error in translation/rotation ($\bm{t}/\bm{r}$) in cm/degree are reported over all scenes along with accuracy under 5cm/5deg. Subscript denotes different versions of A-SCoRe, D-dense, S-sparse, L-lite, B-big. \textbf{\underline{Bold-underline}} indicates best result while \textbf{bold} indicates second best.} 
\label{7scenes_table}
\centering
        \adjustbox{width=\textwidth}{
		\begin{tabular}{l l | c c | c c | c c | c c | c c | c c | c c | c }
		\toprule
		& \multirow{3}{*}{Method} & \multicolumn{14}{c}{7-Scenes} & \multirow{3}{*}{Accuracy} \\
		 & & \multicolumn{2}{c|}{Chess} & \multicolumn{2}{c|}{Fire} & \multicolumn{2}{c|}{Heads} & \multicolumn{2}{c|}{Office} & \multicolumn{2}{c|}{Pumpkin} & \multicolumn{2}{c|}{Red Kitchen} & \multicolumn{2}{c|}{Stairs} & \\
		& & $\bm{t}$, cm & $\bm{r}$, $^\circ$ & $\bm{t}$, cm & $\bm{r}$, $^\circ$ & $\bm{t}$, cm & $\bm{r}$, $^\circ$ & $\bm{t}$, cm & $\bm{r}$, $^\circ$ & $\bm{t}$, cm & $\bm{r}$, $^\circ$ & $\bm{t}$, cm & $\bm{r}$, $^\circ$ & $\bm{t}$, cm & $\bm{r}$, $^\circ$ & \\
	     \midrule
          \multirow{2}{*}{FM} & HLoc[SP]~\cite{sarlin2019coarse} & 
          \textbf{2.0} & 0.85 & 
          \textbf{2.0} & 0.94 & 
          \textbf{\underline{1.0}} & \textbf{\underline{0.75}} &
          3.0 & 0.92 & 
          5.0 & 1.30 & 
          4.0 & 1.40 & 
          5.0 & 1.47 & 
          73.1 \\
          & & & & & & & & & & & & & & & & \\[-7pt]
          & AS~\cite{sattler2016efficient} & 
          3.0 & 0.87 & 
          \textbf{2.0} & 1.01 & 
          \textbf{1.0} & 0.82 & 
          4.0 & 1.15 & 
          7.0 & 1.69 & 
          5.0 & 1.72 & 
          \textbf{4.0} & 1.01 & 
          68.7 \\
          \midrule
          \multirow{3}{*}{ARP}& PoseNet \cite{kendall2015posenet} & 
          13.0 & 4.50 & 
          27.0 & 11.30 & 
          17.0 & 13.00 & 
          19.0 & 5.60 & 
          26.0 & 4.80 & 
          23.0 & 5.40 & 
          35.0 & 12.40&--  \\
          & & & & & & & & & & & & & & & & \\[-7pt]
          & AttTxf~\cite{shavit2021learning} & 
          11.0 & 4.66 & 
          24.0 & 9.60 & 
          14.0 & 12.19 & 
          17.0 & 5.66 & 
          18.0 & 4.44 & 
          17.0 & 5.94 & 
          26.0 & 8.45&-- \\
          & & & & & & & & & & & & & & & & \\[-7pt]
	     & LENS~\cite{moreau2022lens} & 
         3.0 & 1.30 & 
         10.0 & 3.70 & 
         7.0 & 5.80 & 
         7.0 & 1.90 & 
         8.0 & 2.20 & 
         9.0 & 2.20 & 
         14.0 & 3.60 & -- \\
         \midrule
          
          \multirow{5}{*}{SCR} &DSAC++~\cite{brachmann2018learning} & 
          \textbf{2.0} & \textbf{\underline{0.50}} & 
          \textbf{2.0} & 0.90 & 
          \textbf{\underline{1.0}} & \textbf{0.80} & 
          3.0 & \textbf{\underline{0.70}} & 
          4.0 & 1.10 & 
          4.0 & \textbf{1.10} & 
          9.0 & 2.60 & 
          76.1 \\
          & & & & & & & & & & & & & & & & \\[-7pt]
          & DSAC$^\star$~\cite{brachmann2021visual} & 
          \textbf{2.0} & 1.10 & 
          \textbf{2.0} & 1.24 & 
          \textbf{\underline{1.0}} & 1.82 & 
          3.0 & 1.15 & 
          4.0 & 1.34 & 
          4.0 & 1.68 & 
          \underline{\textbf{3.0}} & 1.16 & 
          85.2 \\
          & & & & & & & & & & & & & & & & \\[-7pt]
          & SCRNet~\cite{li2020hierarchical} & 
          \textbf{2.0} & 0.70 & 
          \textbf{2.0} & 0.90 & 
          \textbf{\underline{1.0}} & \textbf{0.80} & 
          3.0 & 0.90 & 
          4.0 & 1.10 & 
          5.0 & 1.40 & 
          \textbf{4.0} & \textbf{1.00} & 
          74.7 \\
          & & & & & & & & & & & & & & & & \\[-7pt]
	     & HSCNet~\cite{li2020hierarchical} & 
         \textbf{2.0} & 0.70 & 
         \textbf{2.0} & 0.90 & 
         \textbf{\underline{1.0}} & 0.90 & 
         3.0 & \textbf{0.80} & 
         4.0 & 1.00 & 
         4.0 & 1.20 & 
         \textbf{\underline{3.0}} & \textbf{\underline{0.80}} & 
         84.8 \\
         & & & & & & & & & & & & & & & & \\[-7pt]
         & D2S\cite{bui2024d2s} & 
         \textbf{2.0} & \textbf{0.68} & 
         2.3 & 0.90 & 
         \textbf{1.3} & \textbf{0.80} & 
         \textbf{2.9} & 0.81 & 
         4.0 & 1.04 & 
         4.2 & 1.29 & 
         13 & 2.02 & 
         79.48 \\
         \midrule
         \multirow{3}{*}{Proposed} & A-SCoRe\textsubscript{D-B} & 
         \textbf{\underline{1.9}} & 0.73 & 
         \textbf{\underline{1.9}} & \textbf{\underline{0.80}} & 
         1.5 & 0.9 & 
         3.0 & 0.89 & 
         \underline{\textbf{3.6}} & \underline{\textbf{0.97}} & 
         \textbf{3.7} & 1.16 & 
         \textbf{4.0} & 1.13 & 
         81.9\\
         & & & & & & & & & & & & & & & & \\[-7pt]
         & A-SCoRe\textsubscript{D-L} &  
         \textbf{2.0} & 0.78 & 
         \textbf{2.0} & \textbf{0.83} & 
         1.4 & 1.12 & 
         \textbf{\underline{2.8}} & 0.81 & 
         \textbf{3.8} & \underline{\textbf{0.97}} & 
         \underline{\textbf{3.4}} & \underline{\textbf{1.04}} & 
         4.6 & 1.25 & 80.96 \\
         & & & & & & & & & & & & & & & & \\[-7pt]
         & A-SCoRe\textsubscript{S} & 
         2.2 & 0.75 & 
         3.7 & 1.41 & 
         3.4 & 2.1 & 
         4.7 & 1.2 & 
         4.7 & 1.1 & 
         5.0 & 1.3 & 
         46 & 9.4 & \\
         \bottomrule
		\end{tabular}

}	

\end{table*}


\begin{figure*}[t]  
    \centering
    \includegraphics[scale=1]{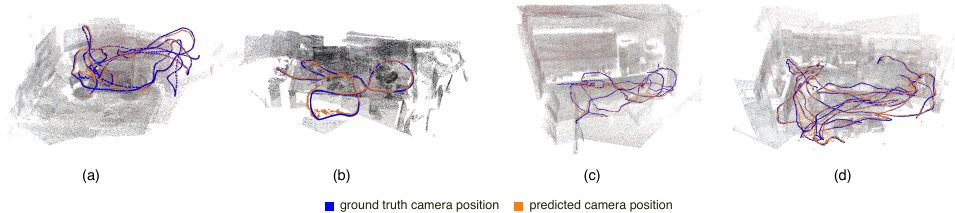} 
    \caption{\textbf{Ground truth camera pose and predicted camera poses comparison}. From (a) to (d) are Chess, Heads, Pumpkin and Redkitchen.} 
    \label{cam_pose_predd}
\end{figure*}

\subsection{Datasets}
Evaluation was done using 2 indoor and 1 outdoor localization datasets, 7Scenes \cite{shotton2013scene}, 12Scenes \cite{valentin2016learning}, and Cambridge Landmarks \cite{kendall2015posenet}.
\begin{itemize}
    \item \textbf{7Scenes} contains seven recordings of RGB-D images collected using KinectFusion, which provided the ground-truth camera poses. This dataset featuring difficult conditions like motion blur, repeated pattern, texture-less regions and reflective surfaces. Each recording has a different number of images for training, ranging from 1k to 7k images. A flaw when collecting this dataset is that the depth channels are not registered to the color images, which can not be used in raw form. We took the depth intrinsics from \cite{brachmann2017dsac} to perform depth calibration and align it to RGB images.
    \item \textbf{12Scenes} is also an indoor dataset similar to 7Scenes. The scenes in 12Scenes are slightly larger than 7Scenes but the number of collected images are lower, which contains several hundred frames. The depth from 12Scenes are registered to color images, which can be used without preprocessing. In addition, a 3D scan for each scene is also available.
    \item \textbf{Cambridge} is a RGB outdoor localization dataset, containing five location collected in Cambridge, UK. Unlike the previous two, cambridge cover a much larger area. It also present challenging conditions such as high illuminations and dynamic objects. For this dataset, we took the SfM 3D model provided from \cite{sarlin2019coarse} as groundtruth.

\end{itemize}

\subsection{Implementation details}
For the image encoder, we experiment with 2 version of CNN backbones with architecture similar to Superpoint \cite{detone2018superpoint}. The first version, dubbed "\textbf{lite}" output dimension $D=256$ and only contains 1.4 millions parameters. The second version, dubbed "\textbf{big}" adds 2 convolution layers, which increase the output channels from 256 to 512.

The feature transformers number of head $h=4$ and number of layer $L=5$. First layer of MLP share the same size as the output dimension of feature transformer, then the following layers dimension are (512, 1024, 1024, 3).

For selection of keypoints, we use pre-trained Superpoint network \cite{detone2018superpoint} due to its robustness. An implementation of PnP RANSAC is borrowed from \cite{brachmann2017dsac} for pose estimation and triangulation given camera poses can be found in COLMAP \cite{schonberger2016structure}.

We used Adam optimizer with $\epsilon=1e-8$,  $\beta \in (0.9, 0.99)$ and a customized learning rate adjustment. With number of total training iteration denoted as $N$, the learning rate at iteration $C$ will be halved and controlled by $lr_C=lr_0\times0.5^k$ where $lr_0$ is the initial learning rate, $\mathbf{k}$ is the decay step $\mathbf{k}=[(C + 2.10^5 - N)/5.10^4]+1$.

\subsection{Indoor results}
We first report the result on 7Scenes dataset in Table \ref{7scenes_table}. We categorized visual localization approaches into 3 groups for easier comparison, namely Feature Matching based (FM), Absolute Pose Regression (ARP), Scene Coordinate Regression (SCR). We present 3 models of A-SCoRe with different model configuration combination. As can be seen, the performance of our A-SCoRe\textsubscript{D-B} is comparable to such other dense approach like DSAC++ \cite{brachmann2018learning}, DSAC* \cite{brachmann2021visual}, HSCNet \cite{li2020hierarchical} and even surpassed them in scenes like Pumpkin, Fire and Redkitchen. D2S\cite{bui2024d2s} work does not based on depth-map but 3D SfM model, which show really good performance on most scenes in 7Scenes. However, using SfM also show a weakness in the case of stairs scene. It achieve poor results with 13 cm/2.02 degree error while our method only has 4cm/1.13 degree, which is 3 times lower than that of D2S in term of translation error. 

\begin{table}[h]
    
\newcolumntype{C}[1]{>{\centering\arraybackslash}p{#1}}
\caption{{\textbf{Storage and number of parameter comparison}. The number of parameters is in the unit of million (M). For Hloc, we use NetVLAD + Superpoint + Superglue configuration, thus we denoted in the table as [SP/SG]}
	}
    \centering
        \renewcommand{\tabcolsep}{1pt}
		\begin{tabular}{C{1cm} | C{1cm} | C{1cm} | C{1cm} | C{1cm} | C{1cm} | C{1cm} | C{1cm}}
		\toprule
        \multirowcell{3}{} & \multicolumn{6}{c}{Methods} \\
        & Hloc [SP/SG] & DSAC++ & SCR & HSCNet & A-SCoRe & A-SCoRe & A-SCoRe\\
        & \cite{sarlin2019coarse} & \cite{brachmann2018learning} & \cite{li2020hierarchical} & \cite{li2020hierarchical} & \textsubscript{D-B} & \textsubscript{D-L} & \textsubscript{S} \\
         \midrule
        No. params & 162.3M & 27M & 25M & 37M & 16.7M & 6.3M & 5.9M \\
         & & & & & & \\ [-7pt]
        Avg. storage & 2.96Gb & 104Mb & 99Mb & 147Mb & 67Mb & 28Mb & 24Mb \\
        \bottomrule
         \end{tabular}

\label{se7scenes_mem}
         
\end{table}

We also provide results of another A-SCoRe version dubbed A-SCoRe\textsubscript{D-L}, which stands for \textbf{dense} and \textbf{lite} in Table \ref{7scenes_table}. Although it is the downsized version of A-SCoRe\textsubscript{D-B}, the performance is still comparable to the \textbf{big} version. Even on specific scene like Office, Pumpkin and Redkitchen, the \textbf{lite} version surpass the \textbf{big} version and achieve the best results among other methods. Only for scene stairs we see the A-SCoRe\textsubscript{D-L} struggle. A-SCoRe\textsubscript{D-L} comes at 4.6 cm and 1.25 degree error while the \textbf{big} version achieve 4.0 cm and 1.13 degree error, which marks the difference of 15\% and 10\% improvements.

Visualization of camera pose prediction from A-SCoRe\textsubscript{D-B} for certain scenes in 7Scenes can be observed in Fig \ref{cam_pose_predd}. As can be seen, for small scale scene like Chess and Heads, the prediction and ground-truth are close with small number of cameras that drift away. On the other hand, bigger scene like Pumpkin and Redkitchen has higher error, thus the camera predictions are also drifted more than the small scene.

The last version of our proposed method, dubbed A-SCoRe\textsubscript{S}, which stands for sparse case is also reported and compared. In real scenarios where the robot is equipped only with RGB camera, our sparse A-SCoRe version would still works at a decent accuracy. As can be seen in Table \ref{7scenes_table}, the translation error and rotation error of sparse version is higher than both dense versions. Scene chess show relatively good performance with 2.2cm/0.75deg error while on the other hand stairs case exhibit 46cm/9.4deg error, which is even higher than absolute pose regression approaches. One main reason for this high error can be seen in Fig \ref{depth_vs_sfm}, where SfM essentially strip down the scene information in textureless areas, which is more common in stairs scene than other scene. This cause lack of information and keypoints to regress, which greatly reduce our model performance.


In Table \ref{se7scenes_mem}, we show the comparison of different approach in term of model weight and storage requirement. For Hloc \cite{sarlin2019coarse}, we calculate the storage as the combination of models (NetVLAD + Superpoint + Superglue) plus the 3D model and descriptor of keypoints, then taking average for all scene in 7Scenes. Because of additional storage for 3D model and descriptor, Hloc takes an significant amount of memory with an average of 2.96Gb. With SCR approaches, since they embbeded the map into the model weight, we only need to account for the storage needed for the model. For comparison, HSCNet has the highest number of parameters with 37M, then followed by DSAC++ and SCRNet. Our model variant A-SCoRe\textsubscript{D-B} and A-SCoRe\textsubscript{D-L} has 16.7M and 6.3M, thus only consume 67Mb and 28Mb of storage. A-SCoRe\textsubscript{D-L} although smaller than HSCNet\cite{li2020hierarchical} by 6 times but results on indoor dataset in Table \ref{7scenes_table} show comparable and even surpassing performance. This result highlight robustness and lightweight of our model.

We perform experiment on 12Scenes dataset and report results in Table \ref{table_12S}. Since this dataset provide registrated depth measurement, we only compare to other dense SCR approaches. Methods like SCRNet \cite{li2020hierarchical}, HSCNet \cite{li2020hierarchical}, achieve high accuracy with multiple scenes reaching 100\%. Our method A-SCoRe comes close after with only slight drops. It is worth mention that SCRNet and HSCNet is a much bigger model compared to our, so they allow bigger learning capacity when the scene become larger. It also reflects that our model performance can be affected by the volume of the scene as for cases like Manolis, Floor5a, and Luke are scenes with large volume in the dataset.

\subsection{Outdoor results}
Table \ref{cambridge_table} presents localization results on the Cambridge dataset. Since no ground-truth depth is available, dense methods use rendered depth from \cite{brachmann2017dsac}, though its generation process is not publicly available, posing a drawback.

HLoc outperforms most methods, especially APR and SCR, showcasing its robustness in large-scale scenarios. Among SCR methods, DSAC++ performs best, likely due to its higher parameter count and training paradigm combining regression and reprojection.

We train both \textbf{"dense + lite"} and \textbf{"sparse"} versions. Across all Cambridge scenes, A-SCoRe struggles with Great Court and St. Mary Church, exhibiting higher errors than other SCR methods. While competitive among other SCR in term of storage, it still falls short of structure-based approaches significantly in term of accuracy.

\begin{table}
    \caption{\textbf{12-Scenes localization results}. Similar to the 7-Scenes localization benchmark, we provide the median translation ($\bm{t}$, cm), orientation ($\bm{r}$, $^\circ$) error, and accuracy with the error threshold of $5cm$ and $5^\circ$.
	\label{table_12S}
	}
    \centering
        \renewcommand{\tabcolsep}{1pt}
		\begin{tabular}{l | c c c|  c c c|  c c c| c c c   }
		\toprule
		 \multirowcell{3}{Scenes} & \multicolumn{12}{c}{Methods} \\
		 & \multicolumn{3}{c}{SCRNet\cite{li2020hierarchical}} & \multicolumn{3}{c}{DSAC++\cite{brachmann2018learning}} & \multicolumn{3}{c}{HSCNet~\cite{li2020hierarchical}} & \multicolumn{3}{c}{A-SCoRe\textsubscript{D-B}} \\
		 & $\bm{t}$, cm & $\bm{r}$, $^\circ$ & Acc & $\bm{t}$, cm & $\bm{r}$, $^\circ$ & Acc & $\bm{t}$, cm & $\bm{r}$, $^\circ$ & Acc& $\bm{t}$, cm & $\bm{r}$, $^\circ$ & Acc \\
	     \midrule
         Kitchen-1 & 
         0.8 & {0.4} &{100} & 
         \multicolumn{2}{c}{-}& 100 & 
         0.8 & {0.4} &{100}  & 
         {1.4} & {0.8} &{97.7}  \\
	     Living-1  & 
         1.1 & {0.4} &{100}  & 
         \multicolumn{2}{c}{-}& 100 & 
         1.1 & {0.4} &{100}  & 
         {1.1} & {0.4} &{100}  \\
	     Bed       & 
         1.3& 0.6 &{100}  & 
         \multicolumn{2}{c}{-}& 99.5 & 
         {0.9} & {0.4} &{100}  & 
         1.6 & {0.6} &{99.5}   \\
	     Kitchen-2 & 
         0.8 & 0.4&{100}  & 
         \multicolumn{2}{c}{-}& 99.5 & 
         {0.7} & {0.3} &{100}  & 
         1.0 & 0.5 &{99.5}   \\
	     Living-2  & 
         1.4 & 0.6 &{100}  & 
         \multicolumn{2}{c}{-}& 100 & 
         {1.0} & {0.4} &{100}  & 
         {1.9} & {0.7} &{98}  \\
	     Luke      & 
         2.0 & 0.9 &93.8  & 
         \multicolumn{2}{c}{-}& 95.5 & 
         {1.2} & {0.5} &96.3  & 
         2.6 & 1.0 &{83.5}   \\
	     Gate362   & 
         1.1 & 0.5 &{100}  & 
         \multicolumn{2}{c}{-}& 100 & 
         {1.0} & {0.4} &{100}  & 
         {1.1} & 0.5 &{100}   \\
	     Gate381   & 
         1.6 & 0.7 & 98.8  & 
         \multicolumn{2}{c}{-}& 96.8 & 
         1.2 & 0.6 &{99.1}  & 
         {1.7} & {0.7} & 97  \\
	     Lounge    & 
         1.5 & 0.5 & 99.4 & 
         \multicolumn{2}{c}{-}& 95.1 & 
         1.4 & 0.5 &{100}  & 
         {1.7} & {0.5} &{97.2}  \\
	     Manolis   & 
         1.4&0.7 &97.2   & 
         \multicolumn{2}{c}{-}& 96.4 & 
         {1.1} & {0.5} &{100}  & 
         1.7 & {0.8} &{93}  \\
	     Floor. 5a & 
         1.6 & 0.7 & 97.0   & 
         \multicolumn{2}{c}{-}& 83.7 & 
         {1.2} & {0.5} &{98.8}  & 
         3.2 & {1.39} & 73  \\
	     Floor. 5b  & 
         1.9 &0.6&93.3   & 
         \multicolumn{2}{c}{-}& 95.0 & 
         1.5 & 0.5 &97.3  & 
         {1.4} & {0.4} & {94.8} \\
         \midrule
         Accuracy & \multicolumn{3}{c}{96.4}  & \multicolumn{3}{c}{99.1}&  \multicolumn{3}{c}{99.1} & \multicolumn{3}{c}{{94.4}}  \\
         \bottomrule
		\end{tabular}

\end{table}

\subsection{Ablations}

\begin{table*}[t!]
    
\caption{\textbf{Outdoor localization results on Cambridge}. We report the median translation ($\bm{t}$, cm) and orientation ($\bm{r}$, $^\circ$) error. The best results are in \textbf{bold}.}
\centering
        \renewcommand{\tabcolsep}{5pt}
		\begin{tabular}{l | c c | c c | c c | c c | c c }
		\toprule
		\multirow{3}{*}{Method} & \multicolumn{10}{c}{Cambridge}  \\
		 & \multicolumn{2}{c}{Kings College} & \multicolumn{2}{c}{Great Court} & \multicolumn{2}{c}{Old Hospital} & \multicolumn{2}{c}{Shop Facade} & \multicolumn{2}{c}{St Mary Church}  \\
		 & $\bm{t}$, m & $\bm{r}$, $^\circ$ & $\bm{t}$, m & $\bm{r}$, $^\circ$ & $\bm{t}$, m & $\bm{r}$, $^\circ$ & $\bm{t}$, m & $\bm{r}$, $^\circ$ & $\bm{t}$, m & $\bm{r}$, $^\circ$ \\
	     \midrule
            Posenet~\cite{kendall2015posenet} &
          1.92 & 5.40 &
          3.67 & 6.50 &
          2.31 & 5.38 &
          1.46 & 8.08 &
          2.65 & 8.48 \\
           &  &  &  &  &  &  &  &  &  &  \\ [-7pt]
          AS~\cite{sattler2016efficient} & 
          0.24 & 0.13 & 
          0.13 & 0.22 & 
          0.20 & 0.36 &
          \textbf{4}&0.21& 
          0.08 & 0.25\\
          &  &  &  &  &  &  &  &  &  &  \\ [-7pt]
          HLoc~\cite{sarlin2019coarse} & 
          0.16&\textbf{0.11} &
          \textbf{0.12}&0.20& 
          \textbf{0.15}&\textbf{0.30} &
          \textbf{0.04}&\textbf{0.20}& 
          \textbf{0.07}&\textbf{0.21}\\
          &  &  &  &  &  &  &  &  &  &  \\ [-7pt]
          DSAC++~\cite{brachmann2018learning} & 
          \textbf{13} & 0.40 & 
          0.40 & 0.20 & 
          0.20 & {0.30} & 
          0.06 & 0.30 & 
          0.13 & 0.40 \\
          &  &  &  &  &  &  &  &  &  &  \\ [-7pt]
          DSAC$^\star$~\cite{brachmann2021visual} & 
          0.15 & 0.30 & 
          0.49 & 0.30 & 
          0.21 & 0.40 & 
          0.05 & 0.30 & 
          0.13 & 0.40 \\
          &  &  &  &  &  &  &  &  &  &  \\ [-7pt]
          SCRNet~\cite{li2020hierarchical} &
          0.21 & 0.3 &
          1.25 & 0.6 &
          0.21 & 0.3 &
          0.06 & 0.3 &
          0.16 & 0.5 \\
          &  &  &  &  &  &  &  &  &  &  \\ [-7pt]
          HSCNet~\cite{li2020hierarchical} & 
          0.18 & 0.30 & 
          0.28 & 0.20 & 
          0.19 & \textbf{0.30} & 
          0.06 & 0.30 & 
          0.09 & 0.30 \\
	     \midrule
         A-SCoRe\textsubscript{D-L} &
         0.39 & 0.6 &
         1.64 & 0.9 &
         0.21 & 0.4 &
         0.17 & 0.7 &
         0.70 & 2.15 \\
         &  &  &  &  &  &  &  &  &  &  \\ [-7pt]
         A-SCoRe\textsubscript{S}   &
         0.20 & 0.30 &
         0.95 & 0.47 &
         0.21 & 0.42 &
         0.21 & 0.80 &
         4.02 & 1.45 \\
         \bottomrule
		\end{tabular}
	
	\label{cambridge_table}
\end{table*}

\begin{figure*}[t]  
    \centering
    \includegraphics[scale=0.8]{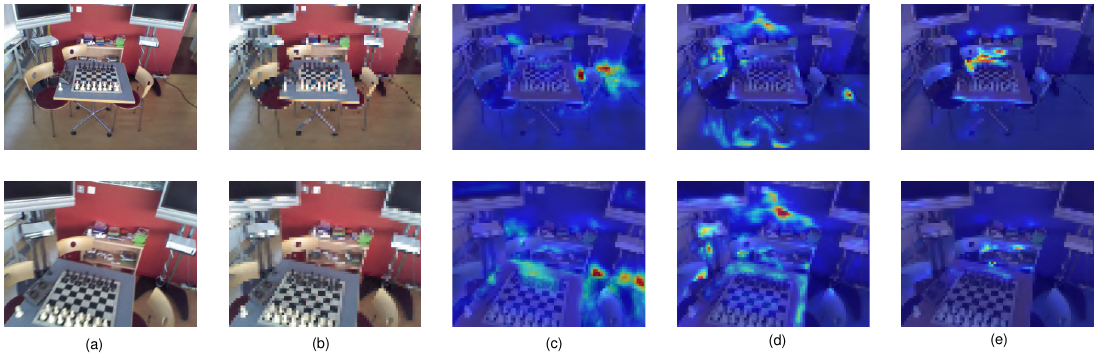} 
    \caption{\textbf{Illustration of attention map as heat map of different heads of the same layer across different images in scene Chess}. (a) and (b) are original image and downsized image after CNN layers. Each heat map from (c) to (e), brighter regions show higher attention score.} 
    \label{attention_map}
\end{figure*}

\textbf{Attention feature map}
The attention map of A-SCoRe\textsubscript{D-B} is illustrated in Fig. \ref{attention_map}. We selected two query images from the Chess scene of the 7Scenes dataset. The heatmaps in columns (c) to (e) are generated by aggregating attention scores across all queries within a single head and then averaging them. In Fig. \ref{attention_map}, we visualize three attention heads of the A-SCoRe\textsubscript{D-B} model.

As seen in the figure, each head focuses on different parts of the image, with warmer regions indicating where the query is concentrated. When the viewpoint changes, the same attention head continues to recognize and focus on specific regions, demonstrating the model's ability to capture semantic information and establish 2D-3D correspondences.

\textbf{Limitations and Future works}
We acknowledge certain limitations in the proposed approach. Although A-SCoRe can work with sparse 3D reconstructions, its reliance on SfM restricts its capability, and there is no fail-safe mechanism in case of SfM failure. Integrating recent deep-learning-based reconstruction methods, such as DUST3R \cite{wang2024dust3r}, could help address this issue.

Our training time is higher than ACE \cite{brachmann2023accelerated} due to the trainable CNN and feature transformer. Thus, potential future work includes exploring pre-training strategies and leveraging efficient transformers like FlashAttention and Linear Attention for speed-up.


\section{CONCLUSIONS}

In this paper, we present A-SCoRe, a new scene coordinate regression architecture that leverages the attention mechanism to incorporate spatial relationships between pixels. Our model can be trained using both dense depth maps and sparse keypoint maps, addressing a limitation of previous models and making it highly versatile across different scenarios and data modalities. Additionally, our model is designed to be scalable, allowing it to be adjusted based on specific requirements. The lightweight version, A-SCoRe\textsubscript{D-L}, with only 6.3M parameters, achieves strong performance against much larger models in indoor scenes. We acknowledge the limitations of our model in outdoor environments and other challenges discussed in the limitations section, which we aim to address in future work.


\bibliographystyle{IEEEtran}  
\bibliography{references}

\begin{thebibliography}{10}
\providecommand{\url}[1]{#1}
\csname url@samestyle\endcsname
\providecommand{\newblock}{\relax}
\providecommand{\bibinfo}[2]{#2}
\providecommand{\BIBentrySTDinterwordspacing}{\spaceskip=0pt\relax}
\providecommand{\BIBentryALTinterwordstretchfactor}{4}
\providecommand{\BIBentryALTinterwordspacing}{\spaceskip=\fontdimen2\font plus
\BIBentryALTinterwordstretchfactor\fontdimen3\font minus \fontdimen4\font\relax}
\providecommand{\BIBforeignlanguage}[2]{{%
\expandafter\ifx\csname l@#1\endcsname\relax
\typeout{** WARNING: IEEEtran.bst: No hyphenation pattern has been}%
\typeout{** loaded for the language `#1'. Using the pattern for}%
\typeout{** the default language instead.}%
\else
\language=\csname l@#1\endcsname
\fi
#2}}
\providecommand{\BIBdecl}{\relax}
\BIBdecl

\bibitem{arandjelovic2016netvlad}
R.~Arandjelovic, P.~Gronat, A.~Torii, T.~Pajdla, and J.~Sivic, ``Netvlad: Cnn architecture for weakly supervised place recognition,'' in \emph{Proceedings of the IEEE conference on computer vision and pattern recognition}, 2016, pp. 5297--5307.

\bibitem{sarlin2019coarse}
P.-E. Sarlin, C.~Cadena, R.~Siegwart, and M.~Dymczyk, ``From coarse to fine: Robust hierarchical localization at large scale,'' in \emph{Proceedings of the IEEE/CVF Conference on Computer Vision and Pattern Recognition}, 2019, pp. 12\,716--12\,725.

\bibitem{sarlin2020superglue}
P.-E. Sarlin, D.~DeTone, T.~Malisiewicz, and A.~Rabinovich, ``Superglue: Learning feature matching with graph neural networks,'' in \emph{Proceedings of the IEEE/CVF conference on computer vision and pattern recognition}, 2020, pp. 4938--4947.

\bibitem{mur2015orb}
R.~Mur-Artal, J.~M.~M. Montiel, and J.~D. Tardos, ``Orb-slam: a versatile and accurate monocular slam system,'' \emph{IEEE transactions on robotics}, vol.~31, no.~5, pp. 1147--1163, 2015.

\bibitem{schonberger2016structure}
J.~L. Schonberger and J.-M. Frahm, ``Structure-from-motion revisited,'' in \emph{Proceedings of the IEEE conference on computer vision and pattern recognition}, 2016, pp. 4104--4113.

\bibitem{kendall2015posenet}
A.~Kendall, M.~Grimes, and R.~Cipolla, ``Posenet: A convolutional network for real-time 6-dof camera relocalization,'' in \emph{Proceedings of the IEEE international conference on computer vision}, 2015, pp. 2938--2946.

\bibitem{bach2022featloc}
T.~B. Bach, T.~T. Dinh, and J.-H. Lee, ``Featloc: Absolute pose regressor for indoor 2d sparse features with simplistic view synthesizing,'' \emph{ISPRS Journal of Photogrammetry and Remote Sensing}, vol. 189, pp. 50--62, 2022.

\bibitem{wang2020atloc}
B.~Wang, C.~Chen, C.~X. Lu, P.~Zhao, N.~Trigoni, and A.~Markham, ``Atloc: Attention guided camera localization,'' in \emph{Proceedings of the AAAI Conference on Artificial Intelligence}, vol.~34, no.~06, 2020, pp. 10\,393--10\,401.

\bibitem{shotton2013scene}
J.~Shotton, B.~Glocker, C.~Zach, S.~Izadi, A.~Criminisi, and A.~Fitzgibbon, ``Scene coordinate regression forests for camera relocalization in rgb-d images,'' in \emph{Proceedings of the IEEE conference on computer vision and pattern recognition}, 2013, pp. 2930--2937.

\bibitem{brachmann2017dsac}
E.~Brachmann, A.~Krull, S.~Nowozin, J.~Shotton, F.~Michel, S.~Gumhold, and C.~Rother, ``Dsac-differentiable ransac for camera localization,'' in \emph{Proceedings of the IEEE conference on computer vision and pattern recognition}, 2017, pp. 6684--6692.

\bibitem{brachmann2018learning}
E.~Brachmann and C.~Rother, ``Learning less is more-6d camera localization via 3d surface regression,'' in \emph{Proceedings of the IEEE conference on computer vision and pattern recognition}, 2018, pp. 4654--4662.

\bibitem{brachmann2021visual}
------, ``Visual camera re-localization from rgb and rgb-d images using dsac,'' \emph{IEEE transactions on pattern analysis and machine intelligence}, vol.~44, no.~9, pp. 5847--5865, 2021.

\bibitem{brachmann2023accelerated}
E.~Brachmann, T.~Cavallari, and V.~A. Prisacariu, ``Accelerated coordinate encoding: Learning to relocalize in minutes using rgb and poses,'' in \emph{Proceedings of the IEEE/CVF Conference on Computer Vision and Pattern Recognition}, 2023, pp. 5044--5053.

\bibitem{bui2024d2s}
B.-T. Bui, H.-H. Bui, D.-T. Tran, and J.-H. Lee, ``D2s: Representing sparse descriptors and 3d coordinates for camera relocalization,'' \emph{IEEE Robotics and Automation Letters}, 2024.

\bibitem{kukelova2013real}
Z.~Kukelova, M.~Bujnak, and T.~Pajdla, ``Real-time solution to the absolute pose problem with unknown radial distortion and focal length,'' in \emph{Proceedings of the IEEE International Conference on Computer Vision}, 2013, pp. 2816--2823.

\bibitem{li2020hierarchical}
X.~Li, S.~Wang, Y.~Zhao, J.~Verbeek, and J.~Kannala, ``Hierarchical scene coordinate classification and regression for visual localization,'' in \emph{Proceedings of the IEEE/CVF Conference on Computer Vision and Pattern Recognition}, 2020, pp. 11\,983--11\,992.

\bibitem{do2022learning}
T.~Do, O.~Miksik, J.~DeGol, H.~S. Park, and S.~N. Sinha, ``Learning to detect scene landmarks for camera localization,'' in \emph{Proceedings of the IEEE/CVF Conference on Computer Vision and Pattern Recognition}, 2022, pp. 11\,132--11\,142.

\bibitem{tang2023neumap}
S.~Tang, S.~Tang, A.~Tagliasacchi, P.~Tan, and Y.~Furukawa, ``Neumap: Neural coordinate mapping by auto-transdecoder for camera localization,'' in \emph{Proceedings of the IEEE/CVF Conference on Computer Vision and Pattern Recognition}, 2023, pp. 929--939.

\bibitem{dosovitskiy2020image}
A.~Dosovitskiy, L.~Beyer, A.~Kolesnikov, D.~Weissenborn, X.~Zhai, T.~Unterthiner, M.~Dehghani, M.~Minderer, G.~Heigold, S.~Gelly \emph{et~al.}, ``An image is worth 16x16 words: Transformers for image recognition at scale,'' \emph{arXiv preprint arXiv:2010.11929}, 2020.

\bibitem{nguyen2024focustune}
S.~T. Nguyen, A.~Fontan, M.~Milford, and T.~Fischer, ``Focustune: Tuning visual localization through focus-guided sampling,'' in \emph{Proceedings of the IEEE/CVF Winter Conference on Applications of Computer Vision}, 2024, pp. 3606--3615.

\bibitem{detone2018superpoint}
D.~DeTone, T.~Malisiewicz, and A.~Rabinovich, ``Superpoint: Self-supervised interest point detection and description,'' in \emph{Proceedings of the IEEE conference on computer vision and pattern recognition workshops}, 2018, pp. 224--236.

\bibitem{torii201524}
A.~Torii, R.~Arandjelovic, J.~Sivic, M.~Okutomi, and T.~Pajdla, ``24/7 place recognition by view synthesis,'' in \emph{Proceedings of the IEEE conference on computer vision and pattern recognition}, 2015, pp. 1808--1817.

\bibitem{humenberger2022investigating}
M.~Humenberger, Y.~Cabon, N.~Pion, P.~Weinzaepfel, D.~Lee, N.~Gu{\'e}rin, T.~Sattler, and G.~Csurka, ``Investigating the role of image retrieval for visual localization: An exhaustive benchmark,'' \emph{International Journal of Computer Vision}, vol. 130, no.~7, pp. 1811--1836, 2022.

\bibitem{camposeco2019hybrid}
F.~Camposeco, A.~Cohen, M.~Pollefeys, and T.~Sattler, ``Hybrid scene compression for visual localization,'' in \emph{Proceedings of the IEEE/CVF Conference on Computer Vision and Pattern Recognition}, 2019, pp. 7653--7662.

\bibitem{ali2024boq}
A.~Ali-bey, B.~Chaib-draa, and P.~Gigu{\`e}re, ``Boq: A place is worth a bag of learnable queries,'' in \emph{Proceedings of the IEEE/CVF Conference on Computer Vision and Pattern Recognition}, 2024, pp. 17\,794--17\,803.

\bibitem{humenberger2020robust}
M.~Humenberger, Y.~Cabon, N.~Guerin, J.~Morat, V.~Leroy, J.~Revaud, P.~Rerole, N.~Pion, C.~de~Souza, and G.~Csurka, ``Robust image retrieval-based visual localization using kapture,'' \emph{arXiv preprint arXiv:2007.13867}, 2020.

\bibitem{panek2022meshloc}
V.~Panek, Z.~Kukelova, and T.~Sattler, ``Meshloc: Mesh-based visual localization,'' in \emph{European Conference on Computer Vision}.\hskip 1em plus 0.5em minus 0.4em\relax Springer, 2022, pp. 589--609.

\bibitem{zhou2022geometry}
Q.~Zhou, S.~Agostinho, A.~O{\v{s}}ep, and L.~Leal-Taix{\'e}, ``Is geometry enough for matching in visual localization?'' in \emph{European Conference on Computer Vision}.\hskip 1em plus 0.5em minus 0.4em\relax Springer, 2022, pp. 407--425.

\bibitem{pan2024glomap}
L.~Pan, D.~Barath, M.~Pollefeys, and J.~L. Sch\"{o}nberger, ``{Global Structure-from-Motion Revisited},'' in \emph{European Conference on Computer Vision (ECCV)}, 2024.

\bibitem{turkoglu2021visual}
M.~O. Turkoglu, E.~Brachmann, K.~Schindler, G.~J. Brostow, and A.~Monszpart, ``Visual camera re-localization using graph neural networks and relative pose supervision,'' in \emph{2021 International Conference on 3D Vision (3DV)}.\hskip 1em plus 0.5em minus 0.4em\relax IEEE, 2021, pp. 145--155.

\bibitem{sattler2019understanding}
T.~Sattler, Q.~Zhou, M.~Pollefeys, and L.~Leal-Taixe, ``Understanding the limitations of cnn-based absolute camera pose regression,'' in \emph{Proceedings of the IEEE/CVF conference on computer vision and pattern recognition}, 2019, pp. 3302--3312.

\bibitem{ng2021reassessing}
T.~Ng, A.~Lopez-Rodriguez, V.~Balntas, and K.~Mikolajczyk, ``Reassessing the limitations of cnn methods for camera pose regression,'' \emph{arXiv preprint arXiv:2108.07260}, 2021.

\bibitem{brahmbhatt2018geometry}
S.~Brahmbhatt, J.~Gu, K.~Kim, J.~Hays, and J.~Kautz, ``Geometry-aware learning of maps for camera localization,'' in \emph{Proceedings of the IEEE conference on computer vision and pattern recognition}, 2018, pp. 2616--2625.

\bibitem{kendall2016modelling}
A.~Kendall and R.~Cipolla, ``Modelling uncertainty in deep learning for camera relocalization,'' in \emph{2016 IEEE international conference on Robotics and Automation (ICRA)}.\hskip 1em plus 0.5em minus 0.4em\relax IEEE, 2016, pp. 4762--4769.

\bibitem{chen2024marepo}
S.~Chen, T.~Cavallari, V.~A. Prisacariu, and E.~Brachmann, ``Map-relative pose regression for visual re-localization,'' in \emph{CVPR}, 2024.

\bibitem{10801953}
H.-H. Bui, B.-T. Bui, D.-T. Tran, and J.-H. Lee, ``Leveraging neural radiance field in descriptor synthesis for keypoints scene coordinate regression,'' in \emph{2024 IEEE/RSJ International Conference on Intelligent Robots and Systems (IROS)}, 2024, pp. 5581--5588.

\bibitem{zhai2024splatloc}
H.~Zhai, X.~Zhang, B.~Zhao, H.~Li, Y.~He, Z.~Cui, H.~Bao, and G.~Zhang, ``Splatloc: 3d gaussian splatting-based visual localization for augmented reality,'' \emph{arXiv preprint arXiv:2409.14067}, 2024.

\bibitem{brachmann2024scene}
E.~Brachmann, J.~Wynn, S.~Chen, T.~Cavallari, {\'A}.~Monszpart, D.~Turmukhambetov, and V.~A. Prisacariu, ``Scene coordinate reconstruction: Posing of image collections via incremental learning of a relocalizer,'' in \emph{European Conference on Computer Vision}.\hskip 1em plus 0.5em minus 0.4em\relax Springer, 2024, pp. 421--440.

\bibitem{wang2024dust3r}
S.~Wang, V.~Leroy, Y.~Cabon, B.~Chidlovskii, and J.~Revaud, ``Dust3r: Geometric 3d vision made easy,'' in \emph{Proceedings of the IEEE/CVF Conference on Computer Vision and Pattern Recognition}, 2024, pp. 20\,697--20\,709.

\bibitem{lindenberger2023lightglue}
P.~Lindenberger, P.-E. Sarlin, and M.~Pollefeys, ``Lightglue: Local feature matching at light speed,'' in \emph{Proceedings of the IEEE/CVF International Conference on Computer Vision}, 2023, pp. 17\,627--17\,638.

\bibitem{sattler2016efficient}
T.~Sattler, B.~Leibe, and L.~Kobbelt, ``Efficient \& effective prioritized matching for large-scale image-based localization,'' \emph{IEEE transactions on pattern analysis and machine intelligence}, vol.~39, no.~9, pp. 1744--1756, 2016.

\bibitem{shavit2021learning}
Y.~Shavit, R.~Ferens, and Y.~Keller, ``Learning multi-scene absolute pose regression with transformers,'' in \emph{Proceedings of the IEEE/CVF International Conference on Computer Vision}, 2021, pp. 2733--2742.

\bibitem{moreau2022lens}
A.~Moreau, N.~Piasco, D.~Tsishkou, B.~Stanciulescu, and A.~de~La~Fortelle, ``Lens: Localization enhanced by nerf synthesis,'' in \emph{Conference on Robot Learning}.\hskip 1em plus 0.5em minus 0.4em\relax PMLR, 2022, pp. 1347--1356.

\bibitem{valentin2016learning}
J.~Valentin, A.~Dai, M.~Nie{\ss}ner, P.~Kohli, P.~Torr, S.~Izadi, and C.~Keskin, ``Learning to navigate the energy landscape,'' in \emph{2016 Fourth International Conference on 3D Vision (3DV)}.\hskip 1em plus 0.5em minus 0.4em\relax IEEE, 2016, pp. 323--332.

\end{thebibliography}

\end{document}